# Grammatical Relations of Myanmar Sentences Augmented by Transformation-Based Learning of Function Tagging

Win Win Thant[1], Tin Myat Htwe[2] and Ni Lar Thein[3]

[1] University of Computer Studies
Yangon, Myanmar

[2] Computer Software Department, University of Computer Studies
Yangon, Myanmar

[3] University of Computer Studies
Yangon, Myanmar

**Abstract**
In this paper we describe function tagging using Transformation Based Learning (TBL) for Myanmar that is a method of extensions to the previous statistics-based function tagger. Contextual and lexical rules (developed using TBL) were critical in achieving good results. First, we describe a method for expressing lexical relations in function tagging that statistical function tagging are currently unable to express. Function tagging is the preprocessing step to show grammatical relations of the sentences. Then we use the context free grammar technique to clarify the grammatical relations in Myanmar sentences or to output the parse trees. The grammatical relations are the functional structure of a language. They rely very much on the function tag of the tokens. We augment the grammatical relations of Myanmar sentences with transformation-based learning of function tagging.

## 1. Introduction

Function tagging is the process of marking up each word in a text with a corresponding function tag like Subj, Obj, Tim, Pla etc. based both on its definition, as well as its context [1]. It has been developed using the statistical implementations, linguistic rules and sometimes both. Identifying the function tags in a given text is an important aspect of any Natural Language Application. We apply TBL for function tagging by extending the Naïve Bayesian based function tagging that is proposed in [2]. The number of function tags in a tagger may vary depending on the information one wants to capture. In the sentence below, the function tags are appended at the end of each word with '#'. For example:
သူ#PSubj သည်#SubjP ကျောင်း#PPla သို့#PlaP သွားသည်#Verb

Grammatical relations are the process of analyzing an input sequence in order to determine its grammatical structure with respect to a given grammar. They show the sentence structure of Myanmar language by using function tags of the words in a sentence. We describe a context free grammar (CFG) based grammatical relations for Myanmar sentences. In the simple sentence below, the grammatical relations are appended at the end of each phrase with '#'. For example:
သူသည်#Subj ကျောင်းသို့#Pla သွားသည်#Verb
In the complex sentence below, the grammatical relations are appended at the end of each phrase with '#'. For example:
မိုးရွာ#Verb သောကြောင့်#CCS ကျွန်မ#Subj ဈေးသို့#Pla မသွားပါ#Verb

Function tagging and grammatical relations are the important steps in Myanmar to English machine translation. Statistical natural language processing (NLP) research in Myanmar language can only be given a push by the creation of annotated corpus for Myanmar language. In Myanmar language, the availability of the functional annotated tagged corpus is very less and so most of the techniques suffer due to data sparseness problem. We present a method that extends a pre-existing function tagger. Grammatical relations are augmented with transformation-based learning of function tagging.

## 2. Related Work

We [2] proposed 39 function tags for Myanmar Language and addressed the question of assigning function tags to Myanmar words and used a small functional annotated tagged corpus as the training data. In the task of function tagging, we used the output of morphological analyzer which tagged the function of Myanmar sentences with correct segmentation, POS (part-of-speech) tagging and chunking information. We used Naïve Bayesian statistics to disambiguate the possible function tags of each word in the sentence. We evaluated the performance of function tagging for simple and complex sentences. We concluded our remarks on tagging accuracy by giving examples of some of the most frequent errors. We showed some examples of common error types.

Yong-uk Park and Hyuk-chul Kwon [3] tried to disambiguate for syntactic analysis system by many dependency rules and segmentation. Segmentation is made during parsing. If two adjacent morphemes had no syntactic relations, their syntactic analyzer made new segment between these two morphemes, and found out all possible partial parse trees of that segmentation and combined them into complete parse trees. Also they used adjacent-rule and adverb subcategorization to disambiguate of syntactic analysis. Their syntactic analyzer system used morphemes for the basic unit of parsing. They made all possible partial parse trees on each segmentation process, and tried to combine them into complete parse trees.

Mark-Jan Nederhof and Giorgio Satta[4] considered the problem of parsing non-recursive context-free grammars, i.e., context-free grammars that generateed finite languages and presented two tabular algorithms for these grammars. They presented their parsing algorithm, based on the CYK (Cocke–Younger–Kasami) algorithm and Earley's alogrithm. As parsing CFG (context-free grammar), they have taken a small hand-written grammar of about 100 rules. They have ordered the input grammars by size, according to the number of nonterminals (or the number of nodes in the forest, following the terminology by Langkilde (2000)).

## 3. Myanmar Language

The Myanmar language, Burmese, belongs to the Tibeto-Myanmar language group of the Sino-Tibetan family. It is also morphologically rich and agglutinative language. Myanmar words are postpositionally inflected with various grammatical features.

### 3.1 Grammatical Hierarchy in Myanmar

The grammatical hierarchy is a useful notion of successively included levels of grammatical construction operating within and between grammatical levels of analysis [5]. This hierarchy is generally assumed in this study as a heuristic principle for the purposes of laying a foundational understanding of Burmese grammatical units and constructions. This hierarchy is a compositional hierarchy in which lower levels typically are filler units for the next higher level in the hierarchy (Longacre 1970, Pike and Pike 1982). Table 1 shows the hierarchy from the lowest level to the highest.

Table 1: Grammatical Hierarchy

| Text |
| --- |
| Paragraph |
| Sentence |
| Clause |
| Phrase |
| Word |
| Morpheme |

### 3.2 Sentences of Myanmar Language

There are two kinds of sentences according to the syntactic structure of Myanmar language [6][7]. They are simple sentence (SS) and complex sentence (CS). Fig 1 shows the syntactic structure of Myanmar language.

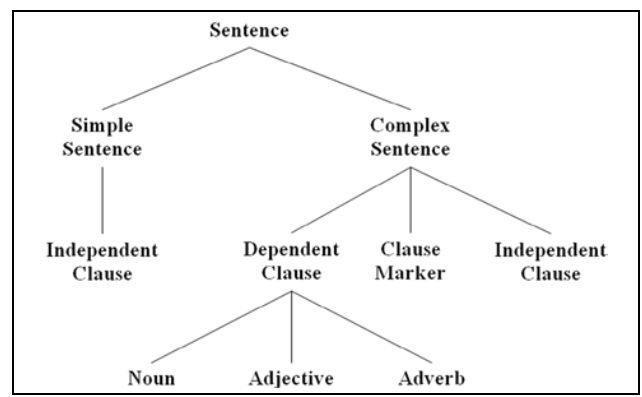

Fig 1: Syntactic Structure

### 3.2.1 Simple Sentence

It contains only one clause. There are two basic phrases such as subject phrase and verb phrase in a simple sentence. For example:

သူ (Subject phrase) အိပ်နေသည်(Verb phrase)

However, a simple sentence can be constructed by only one phrase. This phrase may be verb phrase or noun phrase.
For example:
စားပါ (Verb phrase)
(သူဘယ်သူလဲ) မမ (Noun phrase)

Besides, a simple sentence can be constructed by two or three phrases.
For example:
သွား (Object phrase) တိုက် (Verb phrase)
ရန်ကုန် တွင် (Place phrase) နေသည် (Verb phrase)

Myanmar phrases can be written in any order as long as the verb phrase is at the end of the sentence.
For example:
ဦးဘသည် မန္တလေးမှ ပြန်လာသည်။ (Subject, Place, Verb)
မန္တလေးမှ ဦးဘသည် ပြန်လာသည်။ (Place, Subject, Verb)

A simple sentence can be extended by placing many other phrases between subject phrase and verb phrase. All of the following are simple sentences, because each contains only one clause. It can be quite long.
For example:
ဦးဘသည် ပြန်လာသည်။
U Ba comes back.
ဦးဘသည် မန္တလေးမှ ပြန်လာသည်။
U Ba comes back from Mandalay.
ဦးဘသည် မန္တလေးမှ ရန်ကုန်သို့ ပြန်လာသည်။
U Ba comes back from Mandalay to Yangon.
ဦးဘသည် မန္တလေးမှ ရန်ကုန်သို့ မီးရထားဖြင့် ပြန်လာသည်။
U Ba comes back from Mandalay to Yangon by train.
ဦးဘသည် မန္တလေးမှ ရန်ကုန်သို့ မီးရထားဖြင့် မနက်က ပြန်လာသည်။
U Ba comes back from Mandalay to Yangon by train in the morning.
ဦးဘသည် မောင်မောင်နှင့်အတူ မန္တလေးမှ ရန်ကုန်သို့ မီးရထားဖြင့် မနက်က ပြန်လာသည်။
U Ba comes back from Mandalay to Yangon by train in the morning with Mg Mg.

It is also constructed by adding noun phrases such as subject phrase, object phrase, time phrase and verb phrase. These added noun phrases are called emphatic phrases.
For example:
ပါမောက္ခ ဦးဘသည် သား မောင်မောင်နှင့်အတူ အထက် မန္တလေးမှ မြို့တော် ရန်ကုန်သို့ အမြန် မီးရထားဖြင့် မနေ့ နံနက်က ချောချောမောမော ပြန်လာသည်။
**Professor** U Ba and **his son** Mg Mg came back **safely** from **upper** Mandalay to **capital** Yangon by **express** train in **yesterday** morning.

3.2.2 Complex Sentence

A complex sentence consists of two or more independent clauses (or simple sentences) joined by postpositions, particles or conjunctions. There are at least two verbs or more than two verbs in a complex sentence.

There are two kinds of clause in a complex sentence called independent clause(IC) and dependent clause (DC). DC is in front of IC. A complex sentence contains one independent clause and at least one dependent clause. DC is the same as IC but it must contain a clause marker (CM) in the end. A clause maker may be postpositions, particles or conjunctions [8][9]. There are three dependent clauses depending on the clause marker.

(1)**Noun DC** (joined by postpositions such as မှာ၊ က၊ ကို)
မမ ဈေးသို့ သွားသည် **ကို** ကျွန်မ မြင်သည်။
I see **that** Ma Ma goes to the market.
Noun DC    : မမ ဈေးသို့ သွားသည် **ကို**
IC         : ကျွန်မ မြင်သည်။

(2)**Adjective DC** (joined by particles such as သော ၊ သည်၊ မည့်)
မမ ပေး**သော** စာအုပ် ကို ကျွန်မ ဖတ်သည်။
I read the book **that** is given by Ma Ma.
Adjective DC :မမ ပေး**သော** (စာအုပ်)
IC           :စာအုပ် ကို ကျွန်မ ဖတ်သည်။

(3)**Adverb DC** (joined by conjunctions such as သောကြောင့် ၊ လျက် ၊ သဖြင့်)
မိုးရွာနေ **သောကြောင့်** ကျွန်မဈေးသို့ မသွားပါ။
I do not go to the market **because** it is raining.
Adverb DC    : မိုးရွာနေ **သောကြောင့်**
IC           : ကျွန်မဈေးသို့ မသွားပါ။

## 4. Corpus Creation

Our corpus is to be built manually. We extended the functional annotated tagged corpus that is proposed in [2].We added sentences from newspapers and historical books of Myanmar to the existing corpus. The corpus consists of approximately 5000 sentences with average word length 15 and it is not a balanced corpus that is a bit biased on Myanmar textbooks of middle school. The corpus size is bigger and bigger because the tested sentences are automatically added to the corpus. Myanmar textbooks and historical books are text collections, as shown in Table 2. In our corpus, a sentence contains chunk, function tag, Myanmar word and its POS tag with category. Fig 2 shows the example corpus sentence.

Table 2: Corpus Statistics

| Text types | # of sentences |
|---|---|
| Myanmar textbooks of middle school | 1200 |
| Myanmar grammar books | 700 |
| Myanmar websites | 900 |
| Myanmar newspapers | 750 |
| Myanmar historical books | 1150 |
| Others | 300 |
| Total | 5000 |

```
VC@Active[မိုးရွာ/v.common] #CC@CCS[လျှင်/cc.sent] # NC@Subj
[ကလေး/n.person,များ/part.number] # NC@PPla[လမ်း/n.location] #
PPC@PlaP[ပေါ်တွင်/ppm.place] # NC@Obj[ဘောလုံး/n.objects] #
VC@Active[ကန်ကြ/verb.common]#
SFC@Null[သည်/sf.declarative]။
```

Fig 2: A sentence in the corpus

## 5. Function Tagging by Transformation Based Learning

Transformation-based learning starts with a supervised training corpus that specifies the correct values for some linguistic feature of interest, a baseline heuristics for predicting the values for that feature, and a set of rule templates that determine a space of possible features in the neighborhood surrounding a word, and their action is to change the system's current guess as to the feature for the word. The lexical and the contextual rules are generated from the training corpus [10].

We are not concerned with finding the correct attachment of prepositional phrases. We have stressed at several points that the Naive Bayesian assumptions are crude for many properties of natural language syntax. We describe a method for expressing lexical relations in function tagging that statistical function tagging [2] are currently unable to express. One of the strengths of this method is that it can exploit a wider range of lexical and syntactic regularities. In particular, tags can be conditioned on words and on more contexts. Transformation-based tagging encodes complex interdependencies between words and tags by selecting and sequencing transformations that transform an initial imperfect tagging into one with fewer errors [11]. The training of a transformation-based tagger requires an order of magnitude fewer decisions than estimating the large number of parameters of a Naïve Bayesian model.
A transformation consists of two parts, a triggering environment and a rewrite rule. Table 3 shows examples of the type of transformations that are learned given these triggering environments. The first transformation specifies that Cau should be retagged as PCau when the next tag is "CauP". The first four transformations are triggered by tags and the last three transformations are triggered by words, as shown in Table 3.

Table 3: Examples of some transformations learned in transformation-based tagging

| Source tag | Target tag | Triggering environment |
|---|---|---|
| Cau | PCau | the next tag is CauP |
| PObj | PPla | the second tag is CCC and the fourth tag is PlaP |
| Obj | Subj | the second tag is CCC and the fourth tag is Active |
| Obj | Subj | the second tag is CCC and the fourth tag is CCC and the fifth tag is Active |
| Subj | PcomplS | the lexical item of its next word is "ဖြစ်သည်" |
| Obj | PcomplS | the lexical item of its next word is "နက်သည်" |
| Pla | PcomplS | the lexical item of its next word is "ရှိသည်" |

## 6. Error Analysis for Function Tagging

Transformation rules produced by TBL are then used to change the incorrect tags produced by the Naive Bayesian's method. Interestingly it gave an increase of 0.7% for Myanmar initially the accuracy decreased. This is due to the agglutinative nature of Myanmar and the lack of postpositional marker (PPM) in the sentences. There are about 1200 sentences in the test data for function tagging. Error analysis for function tagging is shown in Table 4.

Table 4: Error Analysis for function tagging

| Actual Tags | Assigned Tags | Counts |
|---|---|---|
| PcomplS | Subj | 133 |
| PcomplS | Obj | 108 |
| PcomplS | Pla | 52 |
| PcomplS | Tim | 24 |
| PSubj | Subj | 28 |
| PObj | Obj | 37 |
| PTim | Tim | 23 |
| PPla | Pla | 18 |
| Subj | Obj | 54 |

## 7. Grammatical Relations

Grammatical functions (or grammatical relations) refer to syntactic relationships between participants in a postposition. Examples are subject, object, time, and place and object complement. We use the context-free grammar (CFG) for grammatical relations of Myanmar sentences. The grammatical relations of the sentences are represented by parse tree. A parse tree is a tree that represents the syntactic structure of a string according to some formal grammar.

The LANGUAGE defined by a CFG is the set of strings derivable from the start symbol S (for Sentence). The core of a CFG grammar is a set of production rules that replaces single variables with strings of variables and symbols. The grammar generates all strings that, starting with a special start variable, can be obtained by applying the production rules until no variables remain. A CFG is usually thought in two ways: a device for generating sentences, or a device if assigning a structure to a given sentence [12]. We use CFG for grammatical relations of function tags.

A CFG is a 4-tuple <N,Σ,P,S> consisting of
- A set of non-terminal symbols N
- A set of terminal symbols Σ
- A set of productions P
  - A-> α
  - A is a non-terminal
  - α is a string of symbols from the infinite set of strings (ΣU N)*
- A designated start symbol S

```
S         → SS| CS
SS        → IC
CS        → Subj? (Noun_DC| Adj_DC| Adv_DC) IC
Noun_DC   → IC CCP
Adj_DC    → IC CCA
Adv_DC    → IC CCS
IC        → Subj Obj Pla Active | Subj Active
Subj      → Subj | PSubj SubjP
Obj       → Obj | PObj  ObjP
Pla       → Pla |  PPla PlaP
Sim       → PSim SimP
Com       → PCom ComP
```

Fig 3: A context free grammar for Myanmar language

## 7.1 Simple Sentence

Consider a simple declarative sentence "သူသည် စာအုပ်ကို ဆရာ့အား ပေးသည်" (He gives the book to the teacher). This sentence is represented as a sequence of function-tags as "PSubj[သူ]# SubjP[သည်]# PObj[စာအုပ်]#ObjP[ကို] #PIobj[ဆရာ့] # IobjP[အား]#Active[ပေးသည်]"

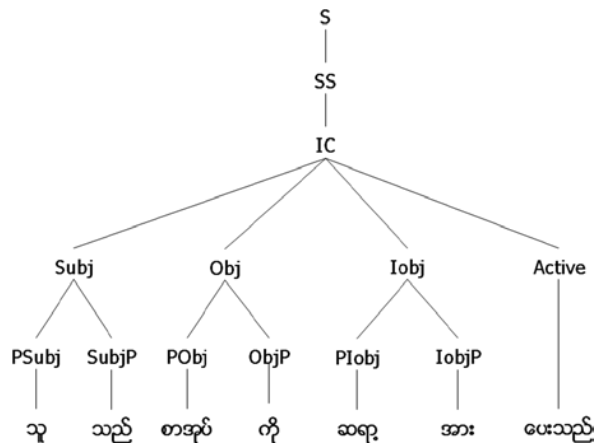

Fig 4: A parse tree for simple sentence

## 7.2 Complex Sentences

### 7.2.1 Complex Sentence joined with postpositions

Consider a complex sentence that is joined with postposition (ကို), "ကလေးများ သစ်ပင်အောက်တွင် ကစားနေသည် ကို ကျွန်တော် မြင်သည်" (I see that children are playing under the tree). This sentence is described as a sequence of function-tags as"Subj[ကလေးများ]#PPla[သစ်ပင်] # PlaP[အောက်တွင်]# Active[ကစားနေသည်]# CCP[ကို]# Subj[ကျွန်တော်]#Active[မြင်သည်]".

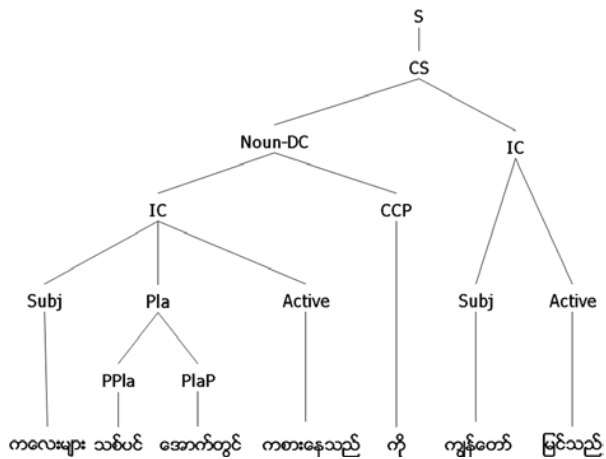

Fig 5: A parse tree for complex sentence (Noun_DC) + (IC)

### 7.2.2 Complex Sentence joined with particles

Consider a complex sentence that is joined with particle (သော), "ကျွန်တော် ဖတ်နေ သော စာအုပ် ကို အဖေ ဝယ်ခဲ့သည်" (I am reading the book that is bought by my father). This sentence is described as a sequence of function-tags as "Subj[ကျွန်တော်]#Active[ဖတ်နေ]#CCA[သော]#PObj[စာအုပ်]#ObjP[ကို]#Subj[အဖေ]#Active[ဝယ်ခဲ့သည်]".

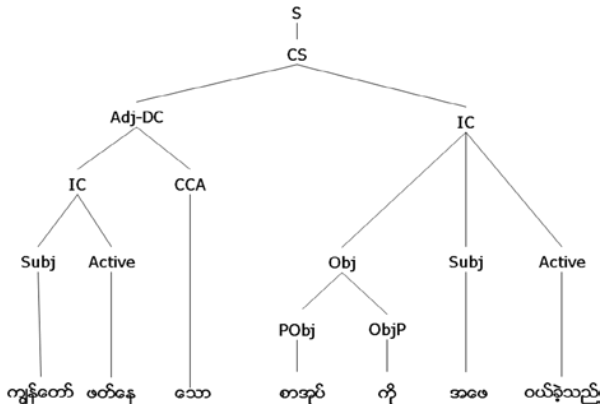

Fig 6: A parse tree for complex sentence (Adj_DC) + (IC)

### 7.2.3 Complex Sentence joined with conjunctions

Consider a complex sentence that is joined with conjunction (သောကြောင့်), "မောင်မောင် ကြိုးစား သောကြောင့် ဂုဏ်ထူး ရသည်" (Mg Mg gets the distinction because he tried). This sentence is represented as a sequence of function-tags as "Subj[မောင်မောင်]#Active[ကြိုးစား]#CCS[သောကြောင့်]#Obj[ဂုဏ်ထူး]#Active[ရသည်]".

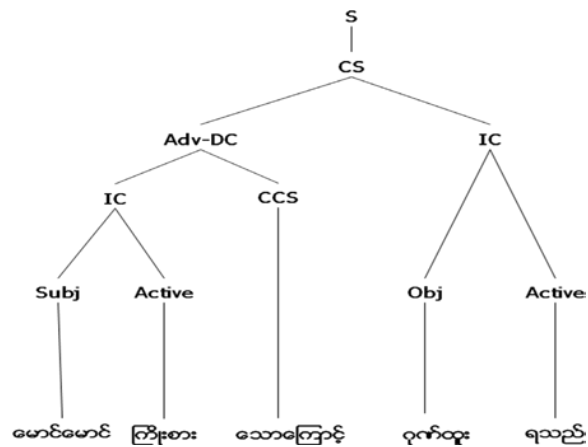

Fig 7: A parse tree for complex sentence (Adv_DC) + (IC)

### 7.2.4 Complicated Complex Sentence

The unrecognized grammatical relations occurs, are the problem that were caused by the DC that are in the middle of IC and do not has a fixed format. DC may exist between the subject phrase and verb phrase of IC. Consider a complex sentence "မောင်ဘ က ကျွန်တော် စာကျက်နေသည် ဟု ပြောသည်" (Mg Ba says that he is studying). This sentence is described as a sequence of function-tags as "PSubj[မောင်ဘ]#SubjP[က]#Subj[ကျွန်တော်]#Active[စာကျက်နေသည်] #CCP[ဟု ]#Active[ပြောသည်]".

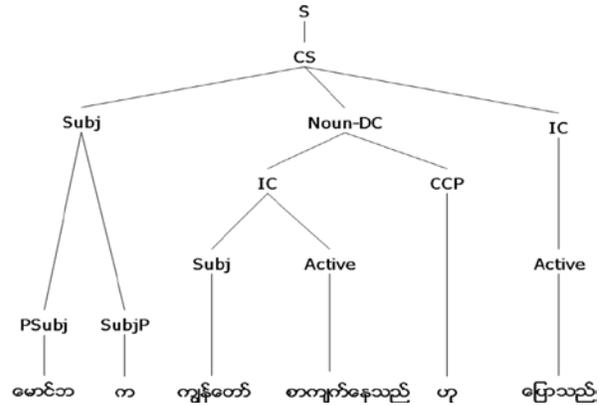

Fig 8: A parse tree for complex sentence Subj+ (Noun_DC) + (IC)

## 8. Performance Evaluation

Evaluation is based on the performance evaluation by comparing between the system's outputs with the manual parse tree of the sentence. By using the way of assessing the quality of grammatical relations is to assign scores to the output sentences. That is affected by POS tagging and function tagging errors. The evaluation steps describe the evaluation methodology:

- Run the system on the selected test case
- Compare the original parse tree with the system's output
- Classify the criteria that arise from the mismatches between the two grammatical relations of the sentences or parse trees
- Assign a suitable score for each criterion. A range of score between 0 and 3 determines the correctness of the relations. While 0 indicates absolutely incorrect grammatical relations and 3 indicates absolutely correct grammatical relations
- When a situation belongs to multiple problems compute its score average
- Determine the correctness of the test case by computing the percentage of the total scores

Table 5: Accuracy scoring criteria

| No | Criterion | Score |
|---|---|---|
| 1 | if the output parse tree is completely wrong format | 0 |
| 2 | if each Myanmar word can generate correct function tag but the grammatical relations are false | 1 |
| 3 | if each Myanmar word cannot generate correct function tag but the grammatical relations are true | 1.5 |
| 4 | if the output sentence is quite well in function tagging and there are some errors in grammatical relations | 2 |
| 5 | if the output parse tree is completely true | 3 |

To the best of our knowledge, there has been no Myanmar-English machine translation before so that there is no standard test set for evaluating Myanmar-English MT system. The data set is derived from the Myanmar textbooks of middle school and Myanmar grammar books, Ministry of Education. The data set consists of 65 sentences for simple sentence, 54 sentences for complex sentence joined with postpositions, 37 sentences for complex sentence joined with particles, 44 sentences for complex sentence joined with conjunctions and 29 sentences for complicated complex sentence.

The system produces 94.36% score for simple sentences while 68.39% score for complicated complex sentences, as shown in Table 6.

Table 6: The result of the score for each sentence type from data set

| No | Sentence Types | No. of sentences | Total Score | Score (%) |
|---|---|---|---|---|
| 1 | Simple | 65 | 184 | 94.36 |
| 2 | Complex (Noun_DC)+(IC) | 54 | 141 | 87.04 |
| 3 | Complex (Adj_DC) +(IC) | 37 | 96.5 | 86.94 |
| 4 | Complex (Adv_DC) + (IC) | 44 | 121 | 91.67 |
| 5 | Complicated Complex | 29 | 59.5 | 68.39 |
|  | Total | 229 | 602 | 87.63 |

Fig 9 depicts the relation accuracy for each sentence type. Table 7 shows detailed expression of the score for each sentence type. It can be seen that the proposed system generates 63.5% accuracy for all sentence types, as shown in Table 7.

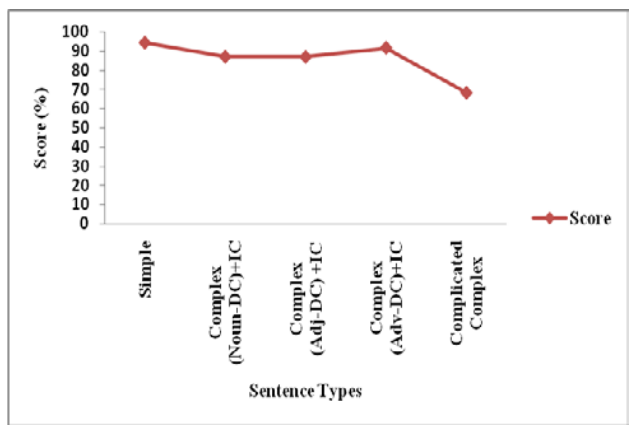

Fig 9: The result of the grammatical relations accuracy for each sentence type

Table 7: The result for each sentence type from the score's point of view

| Sentence Types | Score 3 | Score 2 | Score 1.5 | Score 1 | Score 0 |
|---|---|---|---|---|---|
| Simple | 74.0% | 8.5% | 12.1% | 5.4% | 0.0% |
| Complex (Noun_DC) +(IC) | 67.9% | 0.0% | 29.4% | 2.7% | 0.0% |
| Complex (Adj_DC) +(IC) | 62.2% | 6.3% | 17.4% | 14.1% | 0.0% |
| Complex (Adv_DC) +( IC) | 81.6% | 8.2% | 0.0% | 10.2% | 0.0% |
| Complicated Complex | 32.4% | 1.8% | 19.5% | 46.3% | 12% |
| Accuracy | 63.5% | 4.7% | 15.6% | 15.7% | 2.4% |

Fig 10 to 14 shows the accuracy of grammatical relations for simple and complex sentences. Fig 15 shows the total result of the grammatical relation accuracy from the score point of view.

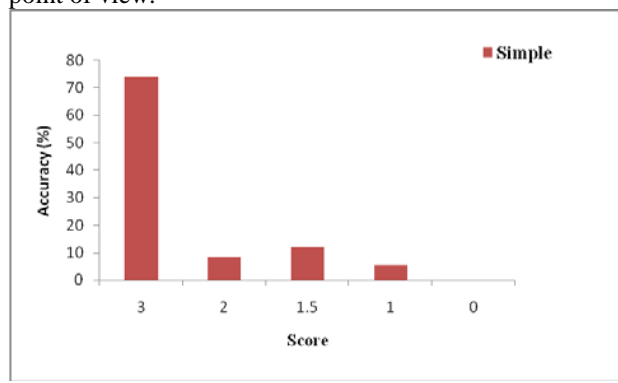

Fig 10: Accuracy for Simple Sentence

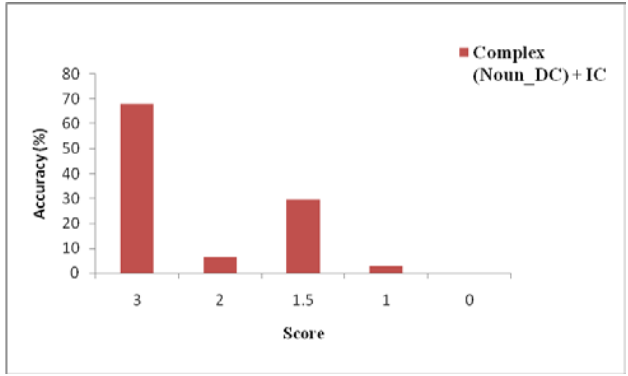

Fig 11: Accuracy for Complex Sentence (Noun_DC) + IC

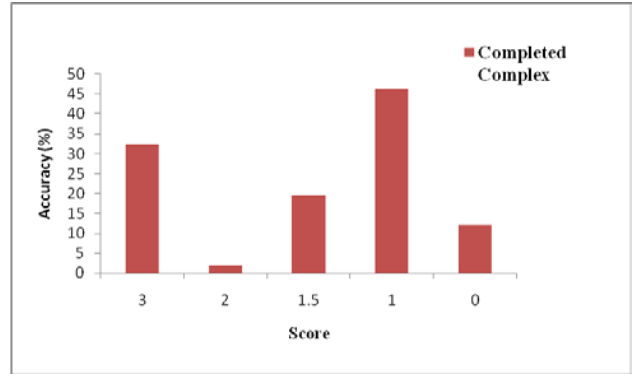

Fig 14: Accuracy for Complicated Complex Sentence

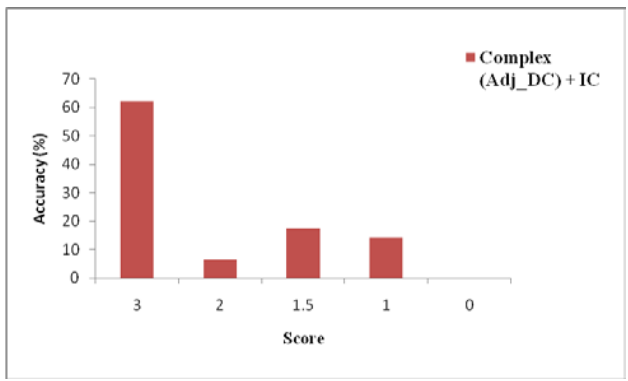

Fig 12: Accuracy for Complex Sentence (Adj_DC) + IC

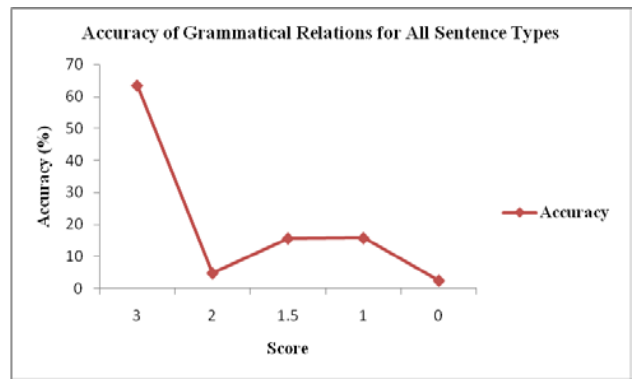

Fig 15: Grammatical relation accuracy for all sentence types from the score point of view

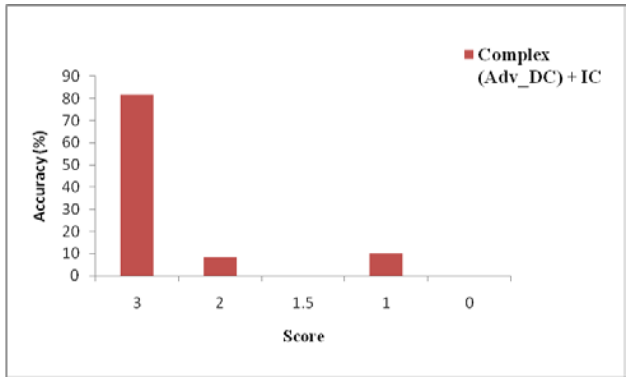

Fig 13: Accuracy for Complex Sentence (Adv_DC) + IC

## 9. Conclusion

We demonstrated the use of TBL for function tagging for Myanmar language. Using TBL method further improved accuracy and produced correct function tags that could not be produced by previous method. Once studied the results and analyzed the mistakes, it must be said that a correct identification of the function tag is crucial in order to obtain a good analysis. If the function tagging fails in this process, the error is dragged throughout the analysis and the result is a badly parse tree. The more accuracy for function tagging increase, the more convenient for grammatical relations of simple sentences and complex sentences of Myanmar language are.

From our experience we have noted that development in natural language processing for Myanmar language is very slow. The main reason for this includes non-availability of large scale data resources and also due to the inherent complexities of the language. The performance of the proposed system can be improved by incorporating more syntactical information by increasing more and more sentence types and well-formed large corpus.

**Appendix**

Table 8: Function Tagset

| Tag | Description | Example |
|---|---|---|
| Active | Verb | စားသည် |
| Subj | Subject | သူ |
| PSubj | Subject | သူ |
| SubjP | PPM of Subject | သည် |
| Obj | Object | ကော်ဖီ |
| PObj | Object | ကော်ဖီ |
| ObjP | PPM of Object | ကို |
| PIobj | Indirect Object | မလှ |
| IobjP | PPM of Indirect Object | အား |
| Pla | Place | ရန်ကုန် |
| PPla | Place | ရန်ကုန် |
| PlaP | PPM of Place | သို့ |
| Tim | Time | မနက် |
| PTim | Time | မနက် |
| TimP | PPM of Time | တွင် |
| PExt | Extract | ကျောင်းသားများ |
| ExtP | PPM of Extract | အနက် |
| PSim | Similie | မင်းသမီး |
| SimP | PPM of Similie | ကဲ့သို့ |
| PCom | Compare | သူ့ထက်လေး |
| ComP | PPM of Compare | နှင့်အတူ |
| POwn | Own | သူ |
| OwnP | PPM of Own | ၏ |
| Ada | Adjective | လှ |
| PcomplS | Subject Complement | သူသည် ဆရာဖြစ် သည် |
| PcomplO | Object Complement | ရွှေကို လက်စွပ်လုပ်သည် |
| PPcomplO | Object Complement | ထွန်းထွန်း |
| PcomplOP | PPM of Object Complement | ဟု |
| PUse | Use | တုတ် |
| UseP | PPM of Use | ဖြင့် |
| PCau | Cause | မိုး |
| CauP | PPM of Cause | ကြောင့် |
| PAim | Aim | အမွေ |
| AimP | PPM of Aim | အတွက် |
| CCS | Join with conjunctions | လျှင် |
| CCM | Join the meanings | ထို့ကြောင့် |
| CCC | Join the words | နှင့် |
| CCP | Join with postpositions | ကို |
| CCA | Join with particles | မည့် |

Table 9: Chunk

| Chunk Type | Example |
|---|---|
| Noun Chunk | NC[ခွေး/n.animal] |
| Postpositional Chunk | PPC[ကို/ppm.obj] |
| Adjectival Chunk | AC[လှ/adj.dem] |
| Adverbial Chunk | RC[ပျော်ရွှင်စွာ/adv.manner] |
| Conjunctional Chunk | CC[နှင့်/cc.chunk] |
| Verb Chunk | VC[ဖြစ်/v.common] |
| Sentence Final Chunk | SFC[၏/sf.declarative] |

Table 10: POS tags

| Description | POS Tag Name |
|---|---|
| Noun | n |
| Pronoun | pron |
| Postpositional Marker | ppm |
| Adjective | adj |
| Adverb | adv |
| Conjunction | cc |
| Particle | part |
| Verb | v |
| Sentence Final | sf |

Table 11: Categories

| Category | Example |
|---|---|
| Noun Categories | n.animal, n.food, n.body, n.person, n.group, n.time, n.common, n.building, n.location, n.objects, n.congnition, |
| Pronoun Categories | pron.person, pron.distplace, pron.disttime, pron.possessive |
| Postpositional Categories | ppm.subj, ppm.obj, ppm.time, ppm.cause, ppm.use, ppm.sim, ppm.aim, ppm.compare, ppm.accept, ppm.place, ppm.extract, |
| Adjectival Categories | adj.dem, adj.distobj |
| Adverbial Categories | adv.manner, adv.state |
| Conjunctional Categories | cc.sent, cc.mean, cc.chunk, cc.part, cc.adj |

| Particle Categories | part.type, part.eg, part.number |
|---|---|
| Verb Categories | v.common, v.compound |
| Sentence Final Categories | sf.declarative, sf.question, sf.negative, |

**Acknowledgments**

We would like to thank Ministry of Science and Technology, Department of Myanmar, Department of English and the Republic of the Union of Myanmar, for promoting a project on Myanmar to English Machine Translation System, where this part of the work was carried out. Large part of this work was carried out at University of Computer Studies, Yangon and our thanks go to all members of the project for their encouragement and support.

**Win Win Thant** is a Ph.D research student. She received B.C.Sc (Bachelor of Computer Science) degree in 2004, B.C.Sc (Hons.) degree in 2005 and M.C.Sc (Master of Computer Science) degree in 2007. She is now an Assistant Lecturer of U.C.S.Y (University of Computer Studies, Yangon). She has written one local paper for Parallel and Soft Computing (PSC) conference in 2010, one international paper for International Conference on Computer Applications (ICCA) conference in 2011 and one journal paper for International Journal of Computer Applications (IJCA) in July 2011. Her research interests include Natural Language Processing and Machine Translation.

**Tin Myat Htwe is** an Associate Professor of U.C.S.Y. She obtained Ph.D degree of Information Technlogy from University of Computer Studies, Yangon. Her research interests include Natural Language Processing, Data Mining and Artificial Intelligence. She has published papers in International conferences and International Journals.

**Ni Lar Thein** is a Rector of U.C.S.Y. She obtained B.Sc. (Chem.), B.Sc. (Hons) and M.Sc. (Computer Science) from Yangon University and Ph.D. (Computer Engg.) from Nanyang Technological University, Singapore in 2003. Her research interests include Software Engineering, Artificial Intelligence and Natural Language Processing. She has published papers in International conferences and International Journals.